\documentclass{article}

\usepackage{microtype}
\usepackage{graphicx}
\usepackage{subfigure}
\usepackage{caption}
\usepackage{booktabs} 
\usepackage[most]{tcolorbox}
\usepackage{hyperref}

\tcbset{
  mybox/.style={
    enhanced,
    colback=gray!10,
    colframe=gray!20!white,
    drop shadow=black!30!white, 
    fonttitle=\bfseries,  
    title=结论,           
    coltitle=black,       
    arc=4mm,              
    boxrule=0.8pt,        
    left=4mm, right=4mm, top=2mm, bottom=2mm, 
  }
}

\usepackage[accepted]{icml2025}

\usepackage{amsmath}
\usepackage{amssymb}
\usepackage{mathtools}
\usepackage{amsthm}
\usepackage{multirow}
\usepackage{bbding}
\usepackage[capitalize,noabbrev]{cleveref}

\theoremstyle{plain}

\theoremstyle{definition}

\theoremstyle{remark}

\usepackage[textsize=tiny]{todonotes}

\icmltitlerunning{Revisit Modality Imbalance at the Decision Layer}

\begin{document}

\twocolumn[
\icmltitle{Revisit Modality Imbalance at the Decision Layer}



\begin{icmlauthorlist}
\icmlauthor{Xiaoyu Ma}{sch1,comp}
\icmlauthor{Hao Chen}{sch1,comp}

\end{icmlauthorlist}

\icmlaffiliation{sch1}{School of Computer Science and Engineering, Southeast University, Nanjing, China}
\icmlaffiliation{comp}{Key Laboratory of New Generation Artificial Intelligence Technology and Its Interdisciplinary
Applications (Southeast University), Ministry of Education, China}

\icmlcorrespondingauthor{Hao Chen}{haochen303@seu.edu.cn}

\icmlkeywords{Balanced Multimodal Learning, Decision Fusion}

\vskip 0.3in
]



\printAffiliationsAndNotice{}  

\begin{abstract}
Multimodal learning integrates information from different modalities to enhance model performance, yet it often suffers from modality imbalance, where dominant modalities overshadow weaker ones during joint optimization. 
This paper reveals that such an imbalance not only occurs during representation learning but also manifests significantly at the decision layer.
Experiments on audio-visual datasets (CREMAD and Kinetic-Sounds) show that even after extensive pretraining and balanced optimization, models still exhibit systematic bias toward certain modalities, such as audio. 
Further analysis demonstrates that this bias originates from intrinsic disparities in feature-space and decision-weight distributions rather than from optimization dynamics alone. 
We argue that aggregating uncalibrated modality outputs at the fusion stage leads to biased decision-layer weighting, hindering weaker modalities from contributing effectively. 
To address this, we propose that future multimodal systems should focus more on incorporate adaptive weight allocation mechanisms at the decision layer, enabling relative balanced according to the capabilities of each modality.
\end{abstract}

\section{Balanced Multimodal Learning}

In an ideal multimodal model, each modality should be fully optimized during training, and the model should be capable of intelligently integrating the contributions of each modality during decision-making. 
However, existing studies have revealed the phenomenon of modality laziness \cite{makes}, where certain modalities fail to be sufficiently learned when the model is optimized with a unified objective. This issue is often attributed to \textit{Modality Imbalance} \cite{ogm} during training. Specifically, due to the Greedy nature of deep models in joint optimization \cite{greedy}, the model tends to prioritize modalities that are easier to learn (the strong modalities), which suppresses the learning of more challenging ones (the weak modalities).

\begin{figure}[t]
\centering
    \subfigure[Heatmap on CREMAD]{\includegraphics[width=0.48\textwidth]{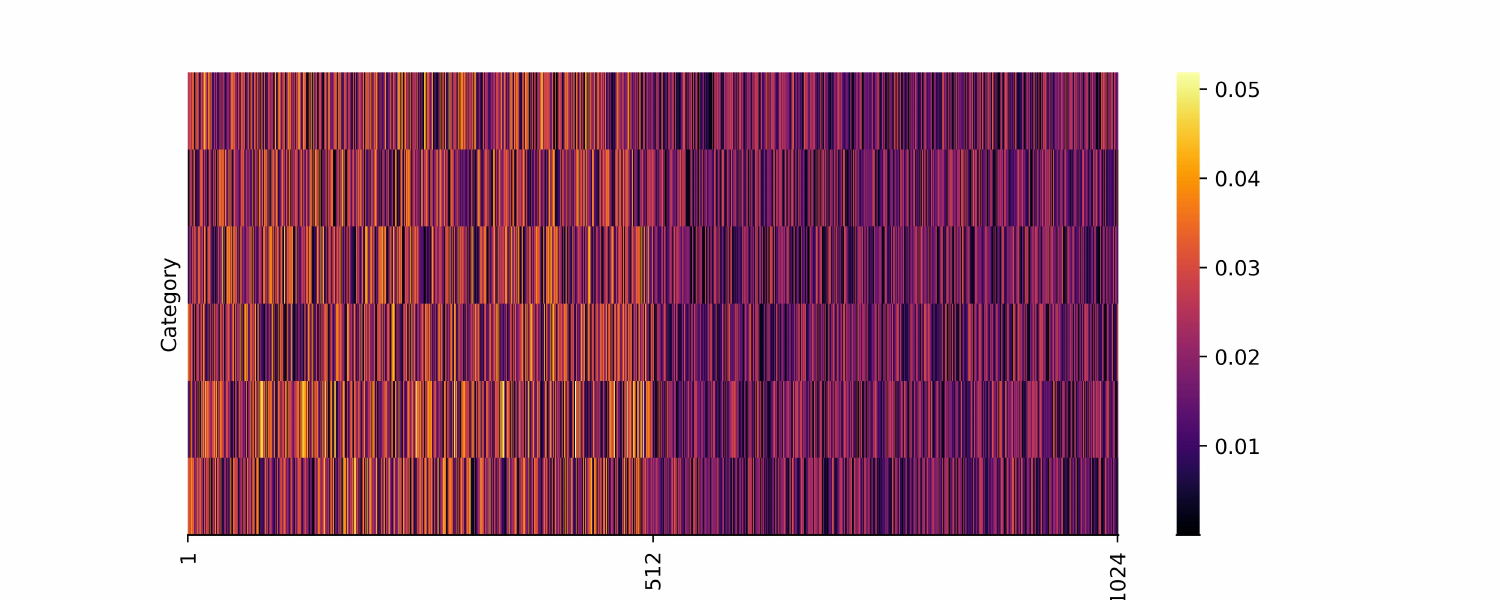}}
    \subfigure[Heatmap on Kinetic-Sounds]{\includegraphics[width=0.48\textwidth]{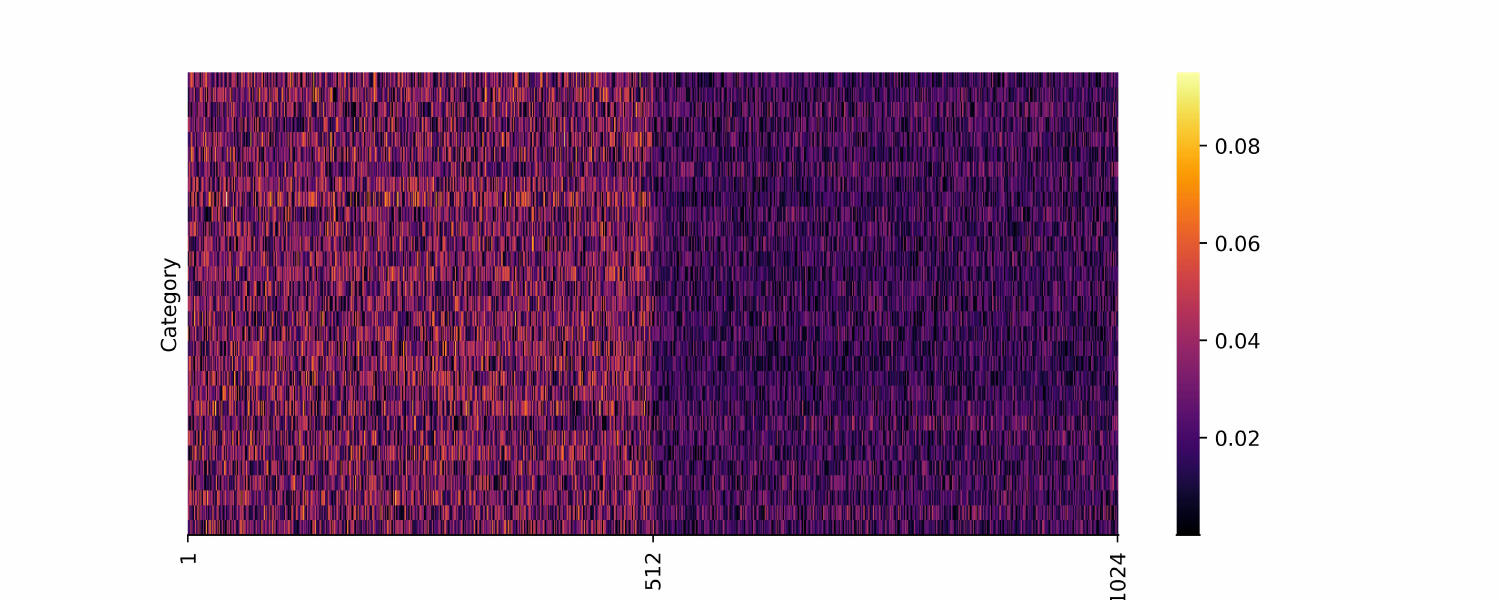}}
\caption{\textbf{Heatmap of decision-layer weights for a joint-trained multimodal model.} The model employs concatenation for feature fusion, with the first 512 dimensions corresponding to audio features and the last 512 dimensions corresponding to video features.}
\label{fig: heatmap}
\end{figure}

To address this problem, research in the field of \textit{Balanced Multimodal Learning(BML)} \cite{bml} has explored various strategies to ensure that all modality encoders are optimized in a more balanced manner. 
Some works mitigate modality laziness by designing modality-specific optimization objectives for the weak modality, such as adjusting task supervision \cite{makes, distillation, pmr, mmpareto}, introducing prototype learning \cite{pmr}, or employing knowledge distillation \cite{distillation}. 
Another line of research aligns optimization dynamics across modalities by adaptively adjusting learning rates \cite{atf, ogm, agm} or gradient updates \cite{mgm, cgm} based on unimodal performance.
Other methods attempt to decouple the multimodal learning process into alternating unimodal learning phases \cite{greedy, mla, remix}.
These methods enhance the learning of individual modality encoders and promote balanced optimization; however, they overlook the fact that \textbf{modality imbalance exists not only in the learning capacity of encoders but also in decision-making}, where the model exhibits significant bias during modality fusion.

\section{Observation of Modality Imbalance at Decision Layer}
In multimodal models, the final decision should be distributed in the fusion layer according to the importance of each modality. To verify this notion, we conducted experiments on two commonly used audio-visual datasets. For the jointly trained multimodal models, we measured the decision-layer weights (L1 norms) and visualized them as heatmaps in \cref{fig: heatmap}. 

The observations reveal a consistent bias across both datasets: the models tend to rely predominantly on features from the audio modality when making final predictions. 
This phenomenon closely resembles the modality imbalance frequently discussed in the balanced multimodal learning literature, suggesting that it represents a decision-level manifestation of modality imbalance.
At the same time, this behavior appears to align with the theoretical expectation that the strong modality with higher predictive performance should receive higher decision weights. But is this truly the case?

\section{Analysis of Modality Imbalance at Decision Layer}
Some researches \cite{mmcosine, online} also observe a phenomenon similar to that shown in \cref{fig: heatmap} and attributed it to modality imbalance arising during the optimization process. \citet{mmcosine} propose MMCosine, a method designed to facilitate the learning of weak modalities by emphasizing the directional alignment of gradients while being independent of their magnitude.
However, can the bias in decision-layer weights truly be explained solely by differences in optimization rates across modalities? After existing BML methods have achieved an optimization balance, do the decision-layer weights also become balanced? Moreover, is a completely balanced decision layer necessarily desirable?
To address these questions, we conducted a series of in-depth experiments.

\begin{figure}[htbp]
\centering
    \subfigure[Heatmap on CREMAD]{\includegraphics[width=0.48\textwidth]{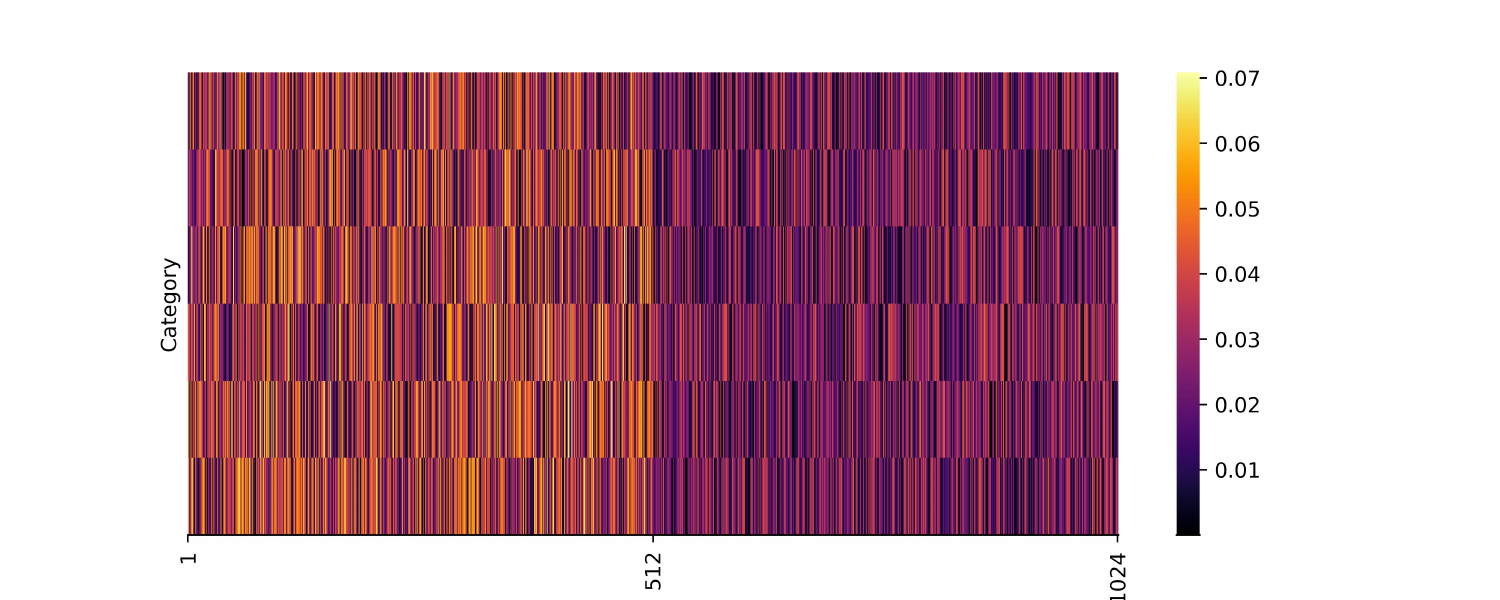}}
    \subfigure[Heatmap on Kinetic-Sounds]{\includegraphics[width=0.48\textwidth]{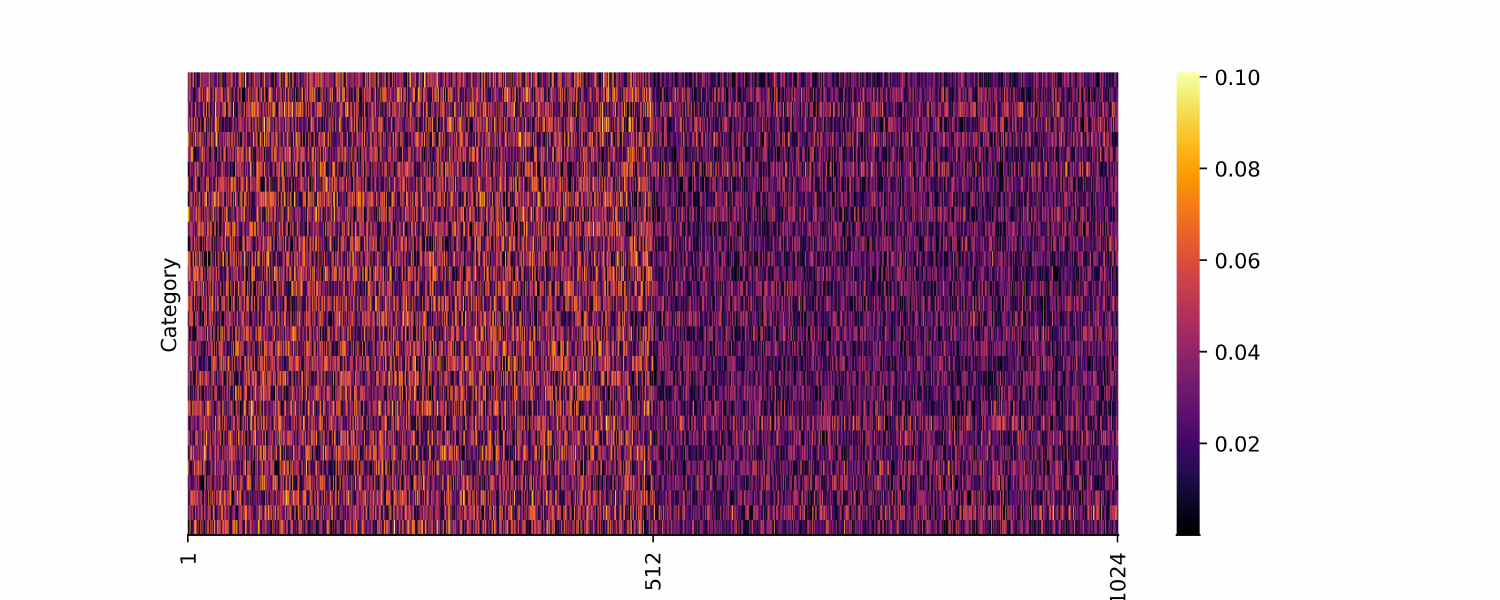}}
\caption{\textbf{Heatmap of decision-layer weights for a pretrain-based multimodal model.} The model employs concatenation for feature fusion, with the first 512 dimensions corresponding to audio features and the last 512 dimensions corresponding to video features.}
\label{fig: heatmap_pretrain}
\end{figure}

\subsection{Possible Causes of Modality Imbalance at the Decision Layer}
Existing studies attribute such modality imbalance at the decision layer to differences in modality optimization rates, where the strong modality dominates the learning process, leading to biased decision-layer weights and logits.
To verify the validity of this explanation, we construct a modality-sufficient multimodal model. Specifically, each modality is first fully pre-trained independently, after which the pre-trained weights are transferred into a multimodal framework. Then, only the decision layer is fine-tuned to achieve feature fusion and final prediction. Although this approach is computationally inefficient, it represents an extreme case that ideally satisfies the expectation of balanced modality learning.

However, as shown in \cref{fig: heatmap_pretrain}, visualizing the decision-layer weights reveals that the modality bias remains significant, even after sufficient pre-training. This observation suggests that attributing the imbalance in decision-layer weights solely to differences in optimization rates is not sufficient, and we get our first insight as follows:

\vspace{0.1in}
\begin{tcolorbox}[mybox, title={Insight 1}]
\textit{Modality imbalance at the decision layer is not merely caused by differences in modality optimization rates, and BML methods that focus on enhancing encoder capabilities cannot address it.}
\end{tcolorbox}
\vspace{0.1in}

\begin{figure}[htbp]
\centering
    \subfigure[Audio - CREMAD]{\includegraphics[width=0.23\textwidth]{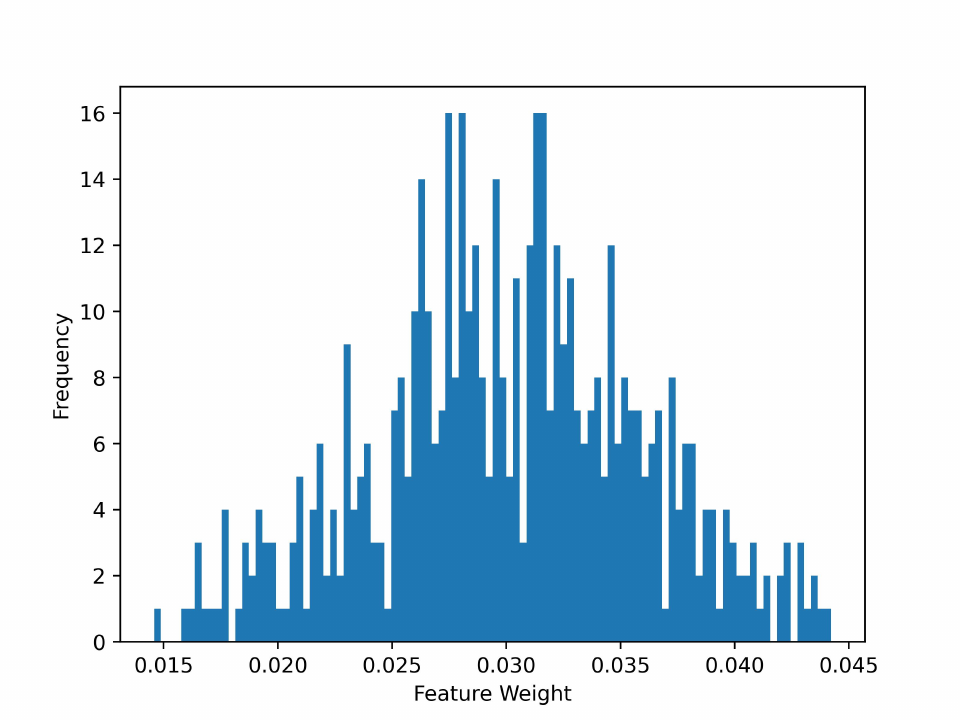}}
    \subfigure[Video - CREMAD]{\includegraphics[width=0.23\textwidth]{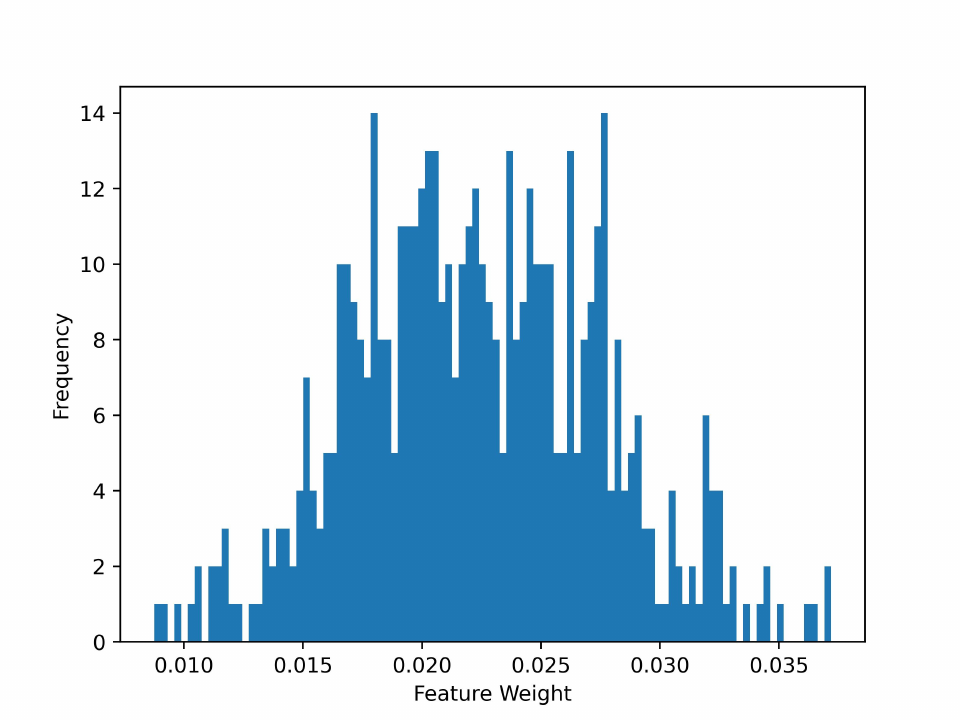}}
    \subfigure[Audio - Kinetic-Sounds]{\includegraphics[width=0.23\textwidth]{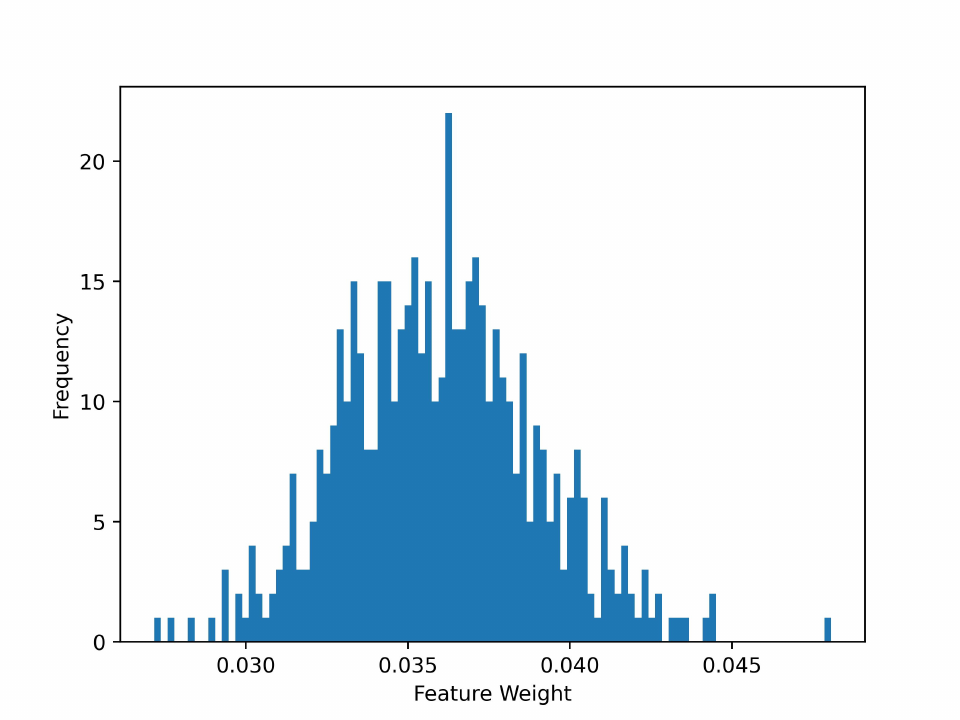}}
    \subfigure[Video - Kinetic-Sounds]{\includegraphics[width=0.23\textwidth]{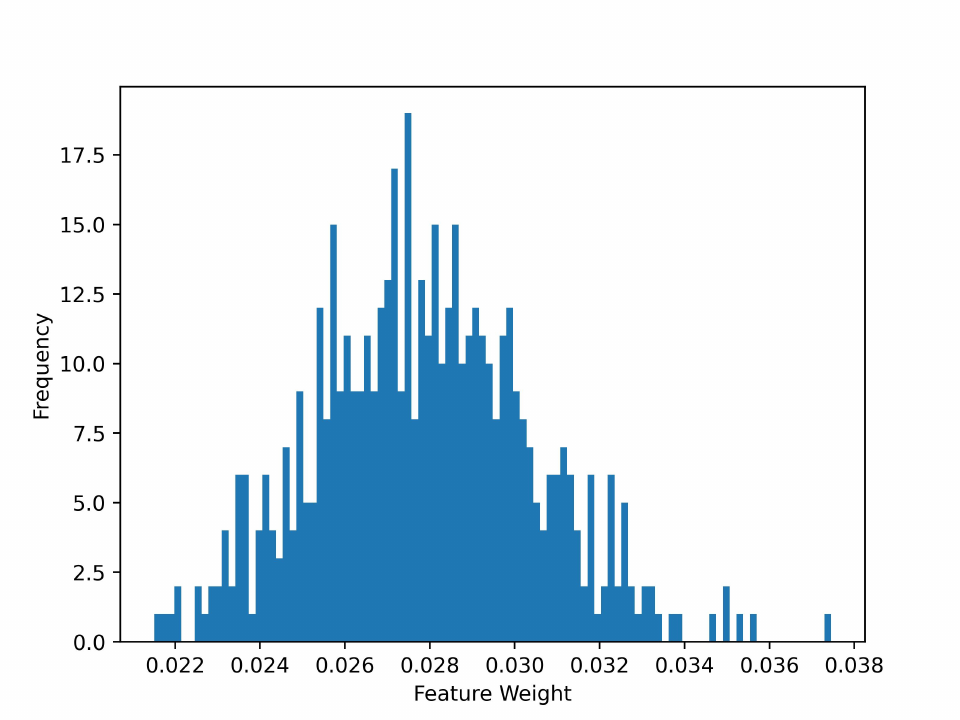}}
\caption{Distribution of decision-layer weights for different unimodal models trained on audiovisual datasets.}
\label{fig: weight_distribution}
\end{figure}

Based on the above results, we hypothesize that the observed bias originates from the inherent differences between modalities themselves. Specifically, due to the discrepancy in feature-space distributions, each modality tends to learn decision weights following its own intrinsic pattern.
To verify this hypothesis, we trained unimodal models separately on both datasets and measured the distribution of their decision-layer weights. As shown in \cref{fig: weight_distribution}, the mean weight magnitude of the audio modality is notably higher than that of the video modality even under unimodal training. This finding supports our assumption that such weight disparities are determined by the intrinsic properties of each modality, rather than being purely an artifact of joint optimization.

\begin{table}[h]
\centering
\begin{tabular}{l|cc||cc}
\toprule
\multirow{2}{*}{\textbf{Method}} & \multicolumn{2}{c||}{\textbf{CREMAD}} & \multicolumn{2}{c}{\textbf{Kinetic-Sounds}} \\
 & Weight & Logits & Weight & Logits \\
\midrule
\textbf{Audio} & 3.56 & 2.14 & 3.63 & 2.47 \\
\midrule
\textbf{Video} & 1.81 & 1.48 & 2.73 & 2.02 \\
\midrule
\textbf{Multi - Audio} & 2.13 & 1.89 & 3.01 & 2.83 \\
\textbf{Multi - Video} & 1.55 & 0.58 & 1.87 & 1.43 \\
\bottomrule
\end{tabular}
\caption{Mean values of Weights $(\times 10^{-2})$ and Logits across modalities on the CREMAD and Kinetic-Sounds datasets.}
\label{tab: weight_logit}
\end{table}

To illustrate this discrepancy more clearly, we summarize in \cref{tab: weight_logit} the mean values of both the decision-layer weights and the output logits. It can be observed that the imbalance in weights further affects the range of the logits. While such range differences have little impact on unimodal prediction due to the subsequent softmax normalization, they can cause bias in the decision layer of multimodal models when concatenation is used for feature fusion. Without proper correction, directly aggregating these imbalanced logits leads to biased decision weights, preventing certain modalities from fully contributing to the final prediction. The above experiments support our second insight:

\vspace{0.1in}
\begin{tcolorbox}[mybox, title={Insight 2}]
\textit{The bias in decision-layer weights originates from the inherent differences in modality data distributions, and the Modality Imbalance at the decision-layer  is an incorrect retention of this phenomenon.}
\end{tcolorbox}
\vspace{0.1in}

\begin{figure}[t]
\centering
    \subfigure[Results on CREMAD]{\includegraphics[width=0.48\textwidth]{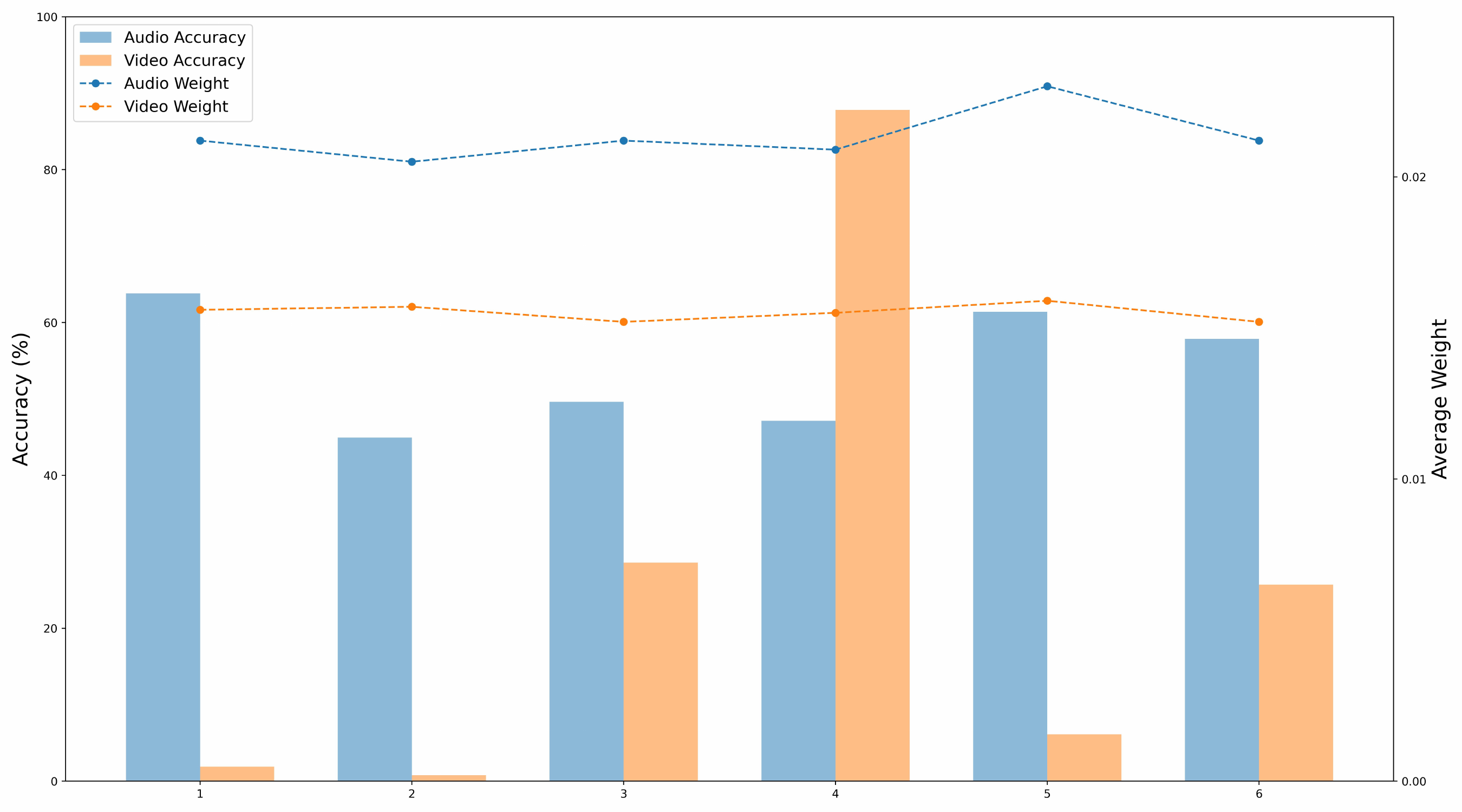}}
    \subfigure[Results on Kinetic-Sounds]{\includegraphics[width=0.48\textwidth]{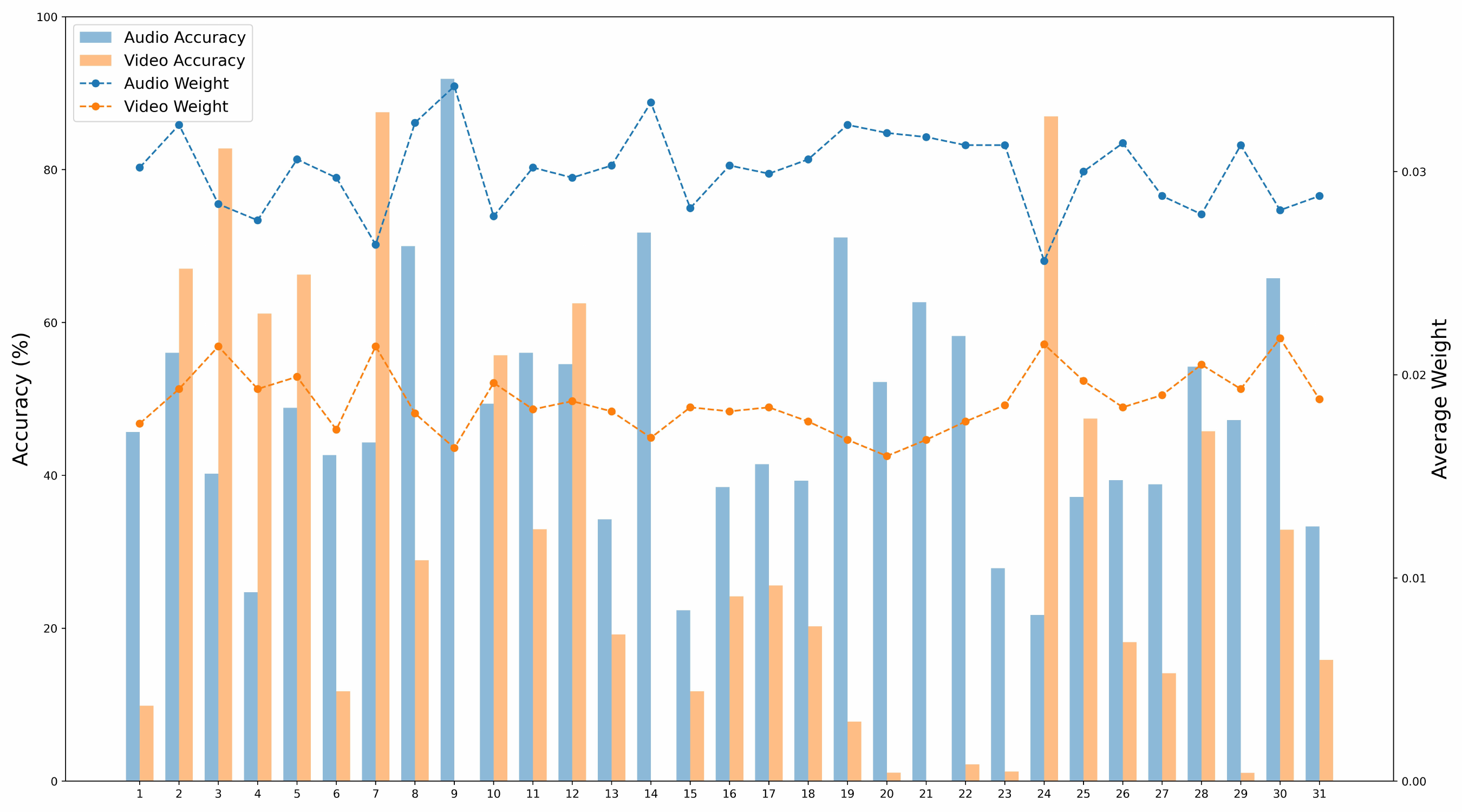}}
\caption{\textbf{Per-Category Accuracy and Modality Weight at the Decision Layer.} The bar chart represents the unimodal prediction accuracy of each modality across categories, while the line chart illustrates the average L1-norm of decision-layer weights for each modality in different categories.}
\label{fig: accuracy_by_cate}
\end{figure}

\subsection{Should We Pursue Decision-Layer Balance?} 
Existing methods enhance encoder capability by promoting balance at the decision layer and logits level. However, such approaches rely on a fundamental assumption that the contribution of each modality at the decision layer aligns with its performance. Yet, does this assumption truly hold?
At the aggregate level, it appears consistent: the audio modality, which dominates the learning process and thus acquires stronger encoding capability, tends to receive higher decision weights.
However, in a classification task, the decision process for each category is relatively independent. Therefore, we further evaluated the modality performance and the average decision-layer weights for each category, as shown in \cref{fig: accuracy_by_cate}.
The results reveal a significant variation in discriminative ability across categories, yet the decision weights consistently exhibit a bias toward the audio modality.
This observation contradicts the common expectation that a multimodal model should assign larger decision weights to the strong modalities according to their predictive capability. Hence, we propose our third insight:

\vspace{0.1in}
\begin{tcolorbox}[mybox, title={Insight 3}]
\textit{Modality Imbalance at the decision layer makes the model cannot automatically allocate decision-layer weights to match the capability of each modality.}
\end{tcolorbox}
\vspace{0.1in}

To address this imbalance, previous studies have primarily focused on promoting alignment at the modality level, thereby emphasizing equal contribution across modalities. However, as shown in \cref{fig: accuracy_by_cate}, different modalities exhibit distinct decision capabilities across categories, and the decision processes are relatively independent and adjustable. Therefore, we argue that such adjustment should occur at the category level and should reflect a capability-aware relative balance, which constitutes our fourth insight:

\vspace{0.1in}
\begin{tcolorbox}[mybox, title={Insight 4}]
\textit{At the decision layer, relative balance should be promoted at the task level (e.g., per category) according to the capabilities of each modality.}
\end{tcolorbox}
\vspace{0.1in}

\section{Conclusion}
In this reports, we argue that modality imbalance at the decision layer is a highly significant yet long-overlooked problem. It arises not only from differences in optimization rates during training but also from inherent disparities in the feature and decision-weight distributions of different modalities. This phenomenon prevents multimodal models from fully leveraging the strengths of each modality, thereby limiting overall performance. Addressing this issue requires more than merely aligning decision-layer weights; it necessitates identifying the modalities that contribute most effectively and optimizing their decision weights accordingly, enabling the model to adaptively adjust weight allocation based on modality capabilities.


\nocite{}

\bibliography{main}
\bibliographystyle{icml2025}

\end{document}